%% file: iclr2026_conference.tex
\pdfoutput=1
\documentclass[11pt]{article}

\usepackage{iclr2027_conference}
\iclrfinalcopy

\usepackage{times}
\usepackage{url}
\usepackage[table]{xcolor}
\usepackage{siunitx}
\definecolor{shadebase}{RGB}{33,102,172}

\definecolor{cascadegray}{gray}{0.93}
\definecolor{oursblue}{RGB}{222,235,247}

\input{math_commands.tex}

\usepackage{latexsym}
\usepackage[T1]{fontenc}
\usepackage{float}
\newfloat{algorithm}{tbp}{loa}
\floatname{algorithm}{Algorithm}

\usepackage[utf8]{inputenc}

\usepackage{microtype}

\usepackage{inconsolata}

\usepackage{graphicx}

\usepackage{amsmath}
\usepackage{cuted}
\usepackage{caption}
\usepackage{booktabs}
\usepackage{placeins}
\usepackage{wrapfig}
\renewcommand{\textfraction}{0.07}

\title{HARISSA: Inference-Time Self-Checks for Efficient and Safe Local Language Model Deployment}

\author{Kenan Alkiek \\
School of Information \\
University of Michigan \\
Ann Arbor, MI 48109, USA \\
\texttt{kalkiek@umich.edu} \\
\And
Moontae Lee \\
SNU-LG AI Research Center \\
Seoul, South Korea \\
\texttt{moontae.lee@lgresearch.ai} \\
\And
David Jurgens \& V.G.Vinod Vydiswaran \\
School of Information \\
University of Michigan \\
Ann Arbor, MI 48109, USA \\
\texttt{\{jurgens,vgvinodv\}@umich.edu} \\
}

\begin{document}
\maketitle
\lhead{Preprint. Under review.}
\begin{abstract}
Running a language model locally offers advantages in privacy, latency, and cost, but local hardware fits only small models, which are less capable than frontier models. The usual remedy for a hard query, escalating it to a cloud model, gives up the privacy and cost advantages of running locally. A deployment that stays local faces two decisions for hard queries instead. First, it can spend more computation on a query, e.g., reasoning before answering, which raises accuracy at a cost in latency, so it must decide which queries are worth the extra computation (efficiency). Second, some queries are beyond the local model, and delivering a wrong answer is worse than deferring the query to a human in the loop, so it must decide which answers are safe to deliver (safety). We show that both decisions can be made from the model's own hidden states. The prefill state, computed before any token is generated, predicts whether the model will answer correctly, and the answer state, at the end of the generated answer, predicts whether that answer is correct. HARISSA fine-tunes the model so that both states predict correctness, then makes both decisions with one policy that cascades through the ways of answering from cheapest to most expensive, skipping a way the prefill state predicts will fail and deferring the query when the answer it stops with is predicted wrong. On a device running a single model, HARISSA is within one accuracy point of chain-of-thought at 2.7 times lower latency. On a server holding four sizes of one model, HARISSA is more accurate than the FrugalGPT and Self-REF cascades at the same latency, and at the same deferral rate the answer state leaves fewer wrong answers than the standard confidence signals in five of six task and setting pairs.
\end{abstract}

\section{Introduction}
\label{sec:intro}

Running a language model locally, for example on a laptop or a phone, avoids the per-token price of a frontier model's cloud API, \$50 per million output tokens.\footnote{Prices as of September 2026: \url{https://platform.claude.com/docs/en/about-claude/pricing} and \url{https://developers.openai.com/api/docs/pricing}.} A local model also answers with no network latency, keeps working offline \citep{xuOnDeviceLanguageModels2024}, and uses less energy per query than a larger model \citep{Luccioni_2024}. A local-only deployment also shares no data with a third party, which any user may value and which some domains require.

Clinical records contain health data, which European data-protection law restricts from transfer outside the Union \citep{gdprRegulation2016} and which locally deployed models can process in place \citep{wiestPrivacypreservingLargeLanguage2024, healthcareSmallLanguageModels2025}. Legal documents contain client confidences, which may not enter a self-learning generative AI tool without the client's informed consent \citep{abaFormalOpinion512GenerativeAI2024}. Financial institutions train and run models on data they hold in-house \citep{wuBloombergGPTLargeLanguage2023}. A frontier model in the cloud can be reached under a vendor contract that covers such data, such as a business associate agreement for health data \citep{openaiDataControls2026}. However, a contract requires legal review, an IT team, and a per-query budget, which many organizations cannot supply. Among hospitals, for example, rural and independent ones adopt AI at lower rates \citep{nongCurrentUseEvaluation2025}, and the rural-health literature names infrastructure, staff, and funding as the barriers \citep{brownGapsArtificialIntelligence2026}.

Consumer hardware such as a laptop or a phone runs only a small language model, currently a few billion parameters \citep{luSmallLanguageModels2024, xuOnDeviceLanguageModels2024}, and the queries a small model answers wrong are the complex ones that need specialized knowledge or multi-step reasoning \citep{dingHybridLLMCostEfficient2024, ongRouteLLMLearningRoute2025}. The standard remedy escalates such a query to the cloud model, either by a router that decides from the query which model answers \citep{ongRouteLLMLearningRoute2025, dingHybridLLMCostEfficient2024} or by a cascade that scores the small model's answer and escalates when the score is low \citep{chenFrugalGPTHowUse2023, yueLargeLanguageModel2024}. For a local-only deployment there is no cloud model to escalate to. A local-only deployment can instead spend more of its own computation on the query \citep{snellScalingLLMTestTime2024}, by reasoning before answering \citep{wei2023chainofthoughtpromptingelicitsreasoning}, by sampling several answers and taking the majority \citep{wang2023selfconsistencyimproveschainthought}, by retrieving from a local corpus \citep{asai2023selfraglearningretrievegenerate, jeong2024adaptiveraglearningadaptretrievalaugmented}, or, on a server, by running a larger size of the model. A local-only deployment can also defer the query to the person \citep{madrasLearningToDefer2018, mozannarConsistentEstimatorsLearning2020, fanconiCascadedLanguageModels2025}. However, more computation raises accuracy at a cost in latency. A 4B model that reasons before answering takes over a minute per query on a data-center GPU (81 s per MedQA query in our experiments, Section~\ref{sec:ondevice}).

A wait of over a minute is prohibitive for a clinician consulting the model between patient visits, a
customer-service agent answering on a live call, or a voice assistant on a phone, yet reasoning often
raises a small model's accuracy on the queries it would otherwise answer wrong, and several recent
small models enable it by default \citep{qwenQwen35_9BModelCard2026, ibmGranite42_8BModelCard2026,
nvidiaNemotron3NanoModelCard2025, huggingfaceSmolLM3ModelCard2025}. A local-only deployment should
therefore answer directly, in a few seconds, and reason only on the queries that need it. Other queries are beyond the model however it answers, and for those a wrong answer delivered is worse than a query deferred, so the deployment should recognize them and defer them to the person. Deciding which queries receive more computation and which are deferred falls to the local model itself, because no larger model is available to judge the query, and a wrong decision either wastes computation or delivers a wrong answer. The signal for each decision is in the model's own hidden states. When the model processes the prompt in one forward pass, the prefill, the hidden state at the
last prompt token predicts whether the model will answer correctly \citep{kadavathLanguageModelsMostly2022a,
lugoloobiLLMsEncodeTheir2026, liangThinkSwitcherWhenThink2025, liuDiffAdaptDifficultyAdaptive2025}, and
the hidden state at the last generated token predicts whether the generated answer is correct
\citep{azariaInternalStateLLM2023, orgadLLMsKnowMore2025}.

We introduce HARISSA (Hidden-Activation Reads for Inference-time Safety and Self-Assessment), a policy that reads the prefill state and the answer state to decide how much computation a query receives and whether its answer is delivered or deferred to the person. HARISSA fine-tunes the model so that each state predicts whether the model's answer is correct, then cascades through the ways of answering from cheapest to most expensive, skipping a way the prefill state predicts will fail and delivering the first answer the answer state predicts correct. As a result, computation goes to the queries that need it, and a query beyond the model reaches the person instead of receiving a wrong answer. The model computes both states in the course of answering, so the decisions add no second model and no generated tokens. Our contributions are threefold. First, on a device running a single model, HARISSA is within one accuracy point of chain-of-thought at 2.7 times lower latency, averaged over three tasks, because the policy reasons only on the queries whose direct answer the answer probe predicts wrong (Section~\ref{sec:ondevice}). Second, on a server holding four sizes of one model, HARISSA is more accurate than FrugalGPT and Self-REF at the same latency, because the prefill probe skips the sizes predicted to fail before anything is generated (Section~\ref{sec:onprem}). Third, deferring by the answer probability leaves fewer wrong answers than deferring by sequence likelihood, the logit margin, or Self-REF confidence at the same deferral rate, on five of six task and setting pairs (Section~\ref{sec:deferral}). All code and data are in the supplementary material and
will be released on publication.

\section{Related Work}
\label{sec:related}

\paragraph{Routing and cascading to a larger model.} Most work on efficient inference chooses which of several models answers a query under a cost target, with a router or with a cascade \citep{wangSurveyCollaboratingSmall2025}. A router picks the model from the query before any model runs. RouteLLM trains one on human preference data to predict which of a strong and a weak model answers better \citep{ongRouteLLMLearningRoute2025}, and Hybrid LLM trains one to predict the quality gap between a small and a large model \citep{dingHybridLLMCostEfficient2024}. A cascade runs the small model first and escalates on a low score. FrugalGPT scores the generated answer with a trained classifier \citep{chenFrugalGPTHowUse2023}, and \citet{guptaLanguageModelCascades2024} learn the escalation rule from the small model's token-level uncertainty. Confident or Seek Stronger escalates from an on-device model to a cloud model when the on-device answer is uncertain \citep{chuangConfidentSeekStronger2025}, and \citet{fanconiCascadedLanguageModels2025} add a human expert after the largest model for the queries every model answered with low confidence. All of these methods rely on a larger model to answer the queries the small model cannot. In a local-only deployment the largest model is still small and may fail, so escalation alone pays the latency of every size and still delivers a wrong answer. HARISSA can skip a size predicted to fail without generating its answer, and defers the query to the person when the answer it stops with is predicted wrong.

\paragraph{Deciding how much computation one model spends, and whether to answer.} Other methods use one model and decide per query how much computation to spend. Sampling several answers and taking the majority raises accuracy at a multiple of the cost \citep{wang2023selfconsistencyimproveschainthought}, and spending that computation only on the queries predicted to need it reaches the same accuracy at a fraction of the cost \citep{snellScalingLLMTestTime2024}. ThinkSwitcher decides from the prefill state whether to reason before answering \citep{liangThinkSwitcherWhenThink2025}, DiffAdapt predicts difficulty from the same state and picks a reasoning strategy to match \citep{liuDiffAdaptDifficultyAdaptive2025}, and AdaptThink retrains the model to choose whether to reason \citep{zhangAdaptThinkReasoningModels2025}. The same kind of estimate also decides whether the model answers at all. Selective prediction withholds the least confident answers and trades coverage for accuracy \citep{elyanivFoundationsNoisefreeSelective2010, geifmanSelectivePrediction2017}, and learning to defer trains a classifier together with a rule that hands a case to a human expert at a cost per deferral \citep{madrasLearningToDefer2018, mozannarConsistentEstimatorsLearning2020}. Each of these methods makes one of the choices a local-only deployment faces, how much to compute or whether to answer, and tunes each on its own. HARISSA makes both decisions with one policy under one objective that prices latency and deferral.

\paragraph{Estimating correctness from a model's own hidden states.} \citet{kadavathLanguageModelsMostly2022a} show that a language model can be trained to predict from the prompt alone whether it knows the answer. \citet{lugoloobiLLMsEncodeTheir2026} show that a linear probe on the prefill state predicts success on mathematics and coding. IntroLM fine-tunes an adapter so that the model predicts during the prefill whether it will answer correctly \citep{kasnaviehIntroLMIntrospectiveLanguage2026}. After generation, \citet{azariaInternalStateLLM2023} classify the truthfulness of a statement from its hidden states, and \citet{orgadLLMsKnowMore2025} find the signal concentrated in the answer tokens and show that such probes transfer poorly across datasets. Self-REF fine-tunes the model to emit a confidence token after its answer \citep{chuangLearningRouteLLMs2025}, and CoKE trains the model to decline the questions its own confidence marks as beyond it \citep{chenTeachingLargeLanguage2024}. An estimate before generation lets a deployment skip an answer that would fail, and an estimate after generation lets the deployment withhold a wrong answer before delivery. Each of these methods provides one estimate or the other. HARISSA fine-tunes one model so that both the prefill state and the answer state predict correctness, and one policy acts on both.

\section{Method}
\label{sec:method}
HARISSA is a policy that cascades through a local deployment's ways of answering a query, from cheapest to most expensive, on a model fine-tuned to predict its own correctness. Before a way of answering generates, a probe on the prefill state decides whether that way is worth running. After a way of answering has generated an answer, a probe on the answer state decides whether to deliver the answer, try the next way, or defer to the person. Fine-tuning the model and fitting the two probes happen once, offline. At inference the policy applies a threshold at each decision, whether to skip a way of answering, whether to stop with an answer or try the next way, and whether to deliver the answer or defer the query.

\paragraph{Problem setting.}\label{sec:setting}
We consider a deployment that can answer a query with any of $K$ \emph{actions}, where an
action is a procedure that takes the query and produces an answer at a measured latency. On a
device the actions are decoding strategies of one model: direct (thinking off) and think
(chain-of-thought reasoning before the answer). On a server the actions are the sizes of one
model family: $2$B, $4$B, $9$B, and $27$B. Actions are indexed $k = 1, \ldots, K$ from cheapest
to most expensive. A person is present and receives any query the deployment does not
answer.

A policy for this deployment takes a query $x$, generates with one or more actions in cost order,
and ends by delivering one answer or by deferring the query to the person. Let $e(x) = 1$ if the
policy delivers a wrong answer, $d(x) = 1$ if it defers, and $T(x)$ be its latency on $x$. The
policy is judged by
\begin{equation}
\mathcal{R} \;=\; \mathbb{E}[e(x)] \;+\; \lambda\,\mathbb{E}[T(x)] \;+\; \mu\,\mathbb{E}[d(x)],
\label{eq:objective}
\end{equation}
where $\lambda$ is the cost of a second of latency and $\mu$ the cost of a deferral, both in
units of one delivered wrong answer; this is the objective of
\citet{fanconiCascadedLanguageModels2025} and \citet{zellingerCostSavingLLMCascades2025}.

The \emph{prefill state} is the hidden state at the last prompt token after the forward pass
over the prompt, so it exists before any token is generated. The \emph{answer state} is the
hidden state at the last token of the generated answer.

\paragraph{Fine-tuning.}\label{sec:finetuning}
Fine-tuning teaches the model to predict, in its own hidden states, whether each action will
answer a query correctly and whether a generated answer is correct. The labels come from the deployment's own training split. Every action answers every training query $x$, each answer is graded, and $y_k(x) \in \{0, 1\}$ records whether action $k$ answered $x$ correctly. We fine-tune the model with LoRA so that its hidden states predict these labels, with linear heads on the two states supplying the training signal. A prefill head for each action $k$ predicts $y_k(x)$ from the prefill state, so the model learns to encode, before generating, which actions will succeed on the query. One answer head predicts $y_k(x)$ from the answer state of action $k$'s answer, so the model learns to encode, after generating, whether the answer it wrote is right. The heads are trained together under a binary cross-entropy loss at equal weight.

\paragraph{Linear probes.}\label{sec:probes}
A probe is a logistic regression on the hidden state at one layer of the fine-tuned model,
calibrated by isotonic regression on the development split
\citep{zadroznyTransformingClassifierScores2002}, and gives the probability that an answer is
correct. For action $k$, the prefill probe reads the
prefill state and gives the \emph{prefill probability} $p_k(x)$, the estimated probability that
action $k$ would answer $x$ correctly; the answer probe reads the answer state and gives the
\emph{answer probability} $q_k(x)$, the estimated probability that the answer generated by
action $k$ is correct. A probe costs one dot product and adds no measurable latency.
Hyperparameters and the probed layer are in Appendix~\ref{app:training}.

\paragraph{The HARISSA policy.}\label{sec:policy}
\begin{wrapfigure}{r}{0.5\textwidth}
\vspace{-\baselineskip}
\small
\begin{tabular}{@{}l@{}}
\texttt{for k = 1, \ldots, K-1 (cheapest first):} \\
\texttt{\ \ \ \ if $p_k(x) < \tau$: continue} \\
\texttt{\ \ \ \ generate action k's answer} \\
\texttt{\ \ \ \ if $q_k(x) \ge \tau_f$: stop with this answer} \\
\texttt{if no answer yet: generate action K's answer} \\
\texttt{deliver the answer if $q \ge \tau_d$, else defer}
\end{tabular}
\captionof{algorithm}{The HARISSA policy. Actions are ordered by cost; $p_k$ and $q_k$ are
the prefill and answer probabilities of action $k$; $\tau$ is the skip threshold, $\tau_f$ the
fall-through threshold, and $\tau_d$ the deferral threshold.}
\label{alg:policy}
\vspace{-\baselineskip}
\end{wrapfigure}
The cascade runs as in Algorithm~\ref{alg:policy}. For each action $k$ but the last, the
policy first applies the \emph{prefill skip}: if $p_k(x)$ is below the \emph{skip threshold}
$\tau$, it skips the action without generating. Otherwise it generates the action's answer and
applies the \emph{answer check}: if $q_k(x)$ is at least the \emph{fall-through threshold}
$\tau_f$, it stops with that answer, and otherwise it discards the answer and moves to action
$k+1$. If the policy reaches the last action, it generates that action's answer and stops with
it, since no action remains to try. The policy then delivers the answer it stopped with,
together with its answer probability, unless that probability is below the \emph{deferral
threshold} $\tau_d$, in which case it withholds the answer and defers the query to the person.
One value of $\tau$ and one of $\tau_f$ apply to every action.

\paragraph{Thresholds.} The skip and fall-through thresholds set how the cascade
trades latency, wrong answers, and deferrals. The skip threshold sets how much latency the
cascade spends: at $\tau = 0$ the policy generates with the cheapest action on every query, as
$\tau$ rises it skips more actions, and above $1$ it runs only the most expensive one. Sweeping
$\tau$ traces an accuracy--latency curve. The skip and fall-through thresholds are chosen on
the development split by minimizing $\mathcal{R}$. We set the deferral threshold equal to the fall-through threshold, so the answer probability that discards an answer at a non-last action also withholds it at the last.

\section{Experimental Setup}
\label{sec:setup}

\paragraph{Datasets and models.}\label{sec:tasks}
We evaluate on three multiple-choice benchmarks: MedQA \citep{jinWhatDiseaseDoes2021}, USMLE-style
clinical questions; MedMCQA \citep{pal2022medmcqalargescalemultisubject}, medical-entrance
questions across subjects; and BBH \citep{suzgun2022challengingbigbenchtaskschainofthought}, general reasoning,
restricted to its sixteen multiple-choice subtasks. Each task has disjoint training,
development, and test splits; split sizes, the BBH subtask list, prompt templates, and
decoding settings are in Appendix~\ref{app:data}. We run every policy in two local-only settings, a device running a single model and a server holding four sizes of one model. On the device we use Qwen~3.5 4B \citep{qwenteamQwen35ModelCard2026, qwen3TechnicalReport2025} with its two actions, direct and think. On the server the actions are the Qwen~3.5 2B, 4B, 9B, and 27B sizes, each answering with thinking off. The device experiments are replicated on two other model families,
Gemma~4 \citep{gemma4TechnicalReport2026} and Cogito, in Appendix~\ref{app:replication}.

\paragraph{Baselines.}\label{sec:eval}
On the device, where a single model answers, we compare HARISSA against three baselines that
answer every query the same way and against two cascades from prior work. The baselines are
direct (thinking off), chain-of-thought (thinking on), and self-consistency, the majority of
three sampled direct answers \citep{wang2023selfconsistencyimproveschainthought}. The cascades
are FrugalGPT and Self-REF. FrugalGPT \citep{chenFrugalGPTHowUse2023} grades each generated
answer with a trained DistilBERT scorer and escalates when the score is below a threshold.
Self-REF \citep{chuangLearningRouteLLMs2025} fine-tunes the model to emit a confidence token
after its answer and escalates on that token's probability. On the server we compare against
the same two cascades. Deferral has its own baselines. A deferral signal ranks the answers a policy delivers, and the lowest-ranked share goes to the person, so at the same share deferred the better signal is the one that removes more of the wrong answers. We compare
the answer probability against three standard confidence signals, each computed on the same
answers HARISSA delivers: sequence likelihood
\citep{malininUncertaintyEstimationAutoregressive2021}, the logit margin (the gap between the
top two option logits at the answer, a variant of the maximum-softmax baseline of
\citealp{hendrycksBaselineDetectingMisclassified2017, geifmanSelectivePrediction2017}), and Self-REF confidence, the probability of the confidence token that the Self-REF model of the same size emits when given the query and HARISSA's answer.

\paragraph{Metrics and protocol.}\label{sec:metrics}
We report accuracy when every query is answered, and, with deferral, wrong answers delivered and the share of queries deferred, each as a percentage of all queries, and mean latency per query in seconds, computed from each query's
token counts at rates measured on one NVIDIA A40 (Appendix~\ref{app:latency}). The model and
the probes are fit on the training split, calibration and the thresholds are chosen on the
development split, and every number we report is on the test split. The skip and fall-through
thresholds are chosen jointly by the objective (Eq.~\ref{eq:objective}), the fall-through
threshold over $0.3$ to $0.7$, with $\lambda$ set so that one accuracy point is worth $10$\,s
of latency per query on the device and $0.5$\,s on the server. Deferral is off except in Section~\ref{sec:deferral}, so that the policies are compared on the same queries at full coverage.

\section{Results}
\label{sec:results}

\begin{figure*}[!t]
\centering
\includegraphics[width=0.8\textwidth]{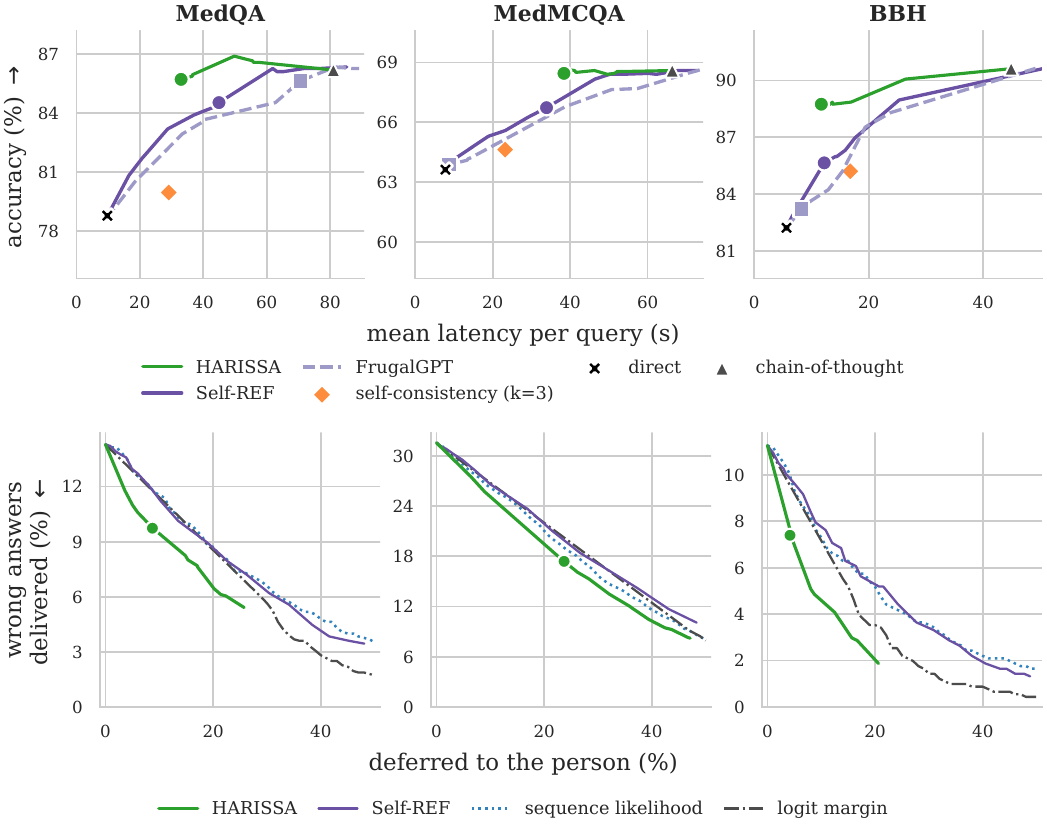}
\caption{\textbf{On a device, HARISSA is within one accuracy point of chain-of-thought at 2.7 times
lower latency, averaged over three tasks.} Top: accuracy against mean latency per query, every query answered; the baselines as markers and each cascade as its sweep over the skip threshold, with the selected point filled. Bottom: wrong answers delivered against the share of queries deferred, on HARISSA's delivered answers, one curve per deferral signal (HARISSA is the answer probability and Self-REF is Self-REF confidence); the filled point is the selected deferral threshold. The answer probability takes few distinct values, so its curve ends early, and answers without a logit margin are deferred first.}
\label{fig:device}
\end{figure*}

\begin{figure*}[!t]
\centering
\includegraphics[width=0.8\textwidth]{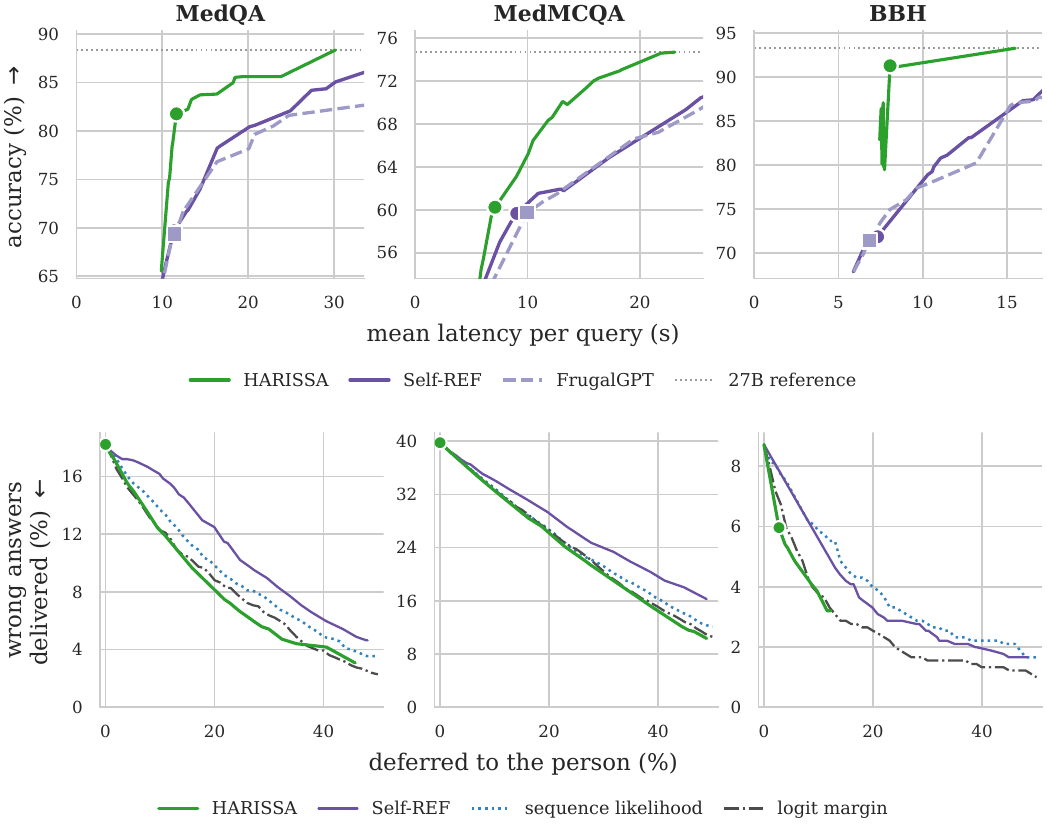}
\caption{\textbf{On a server, HARISSA is more accurate than FrugalGPT and Self-REF at the same latency by skipping the sizes the prefill probe predicts will fail.} Same layout as Figure~\ref{fig:device}; the dotted line (27B reference) is the 27B size alone, the highest accuracy any size or cascade reaches.}
\label{fig:server}
\end{figure*}

HARISSA is more accurate than FrugalGPT and Self-REF at the same latency on the device and on the server (Figures~\ref{fig:device} and~\ref{fig:server}, top; Table~\ref{tab:main}). The answer probability on each delivered answer is also the deferral signal, and it leaves fewer wrong answers than the standard confidence signals at the same deferral rate (Figures~\ref{fig:device} and~\ref{fig:server}, bottom).

\begin{table}[t]
\centering
\small
\caption{\textbf{Accuracy and mean latency per query for each policy at its selected thresholds, every query answered.} Accuracy (Acc.) is in percent and latency (Lat.) in seconds; HARISSA's rows are in bold.}
\label{tab:main}
\begin{tabular}{l
                S[table-format=2.1] S[table-format=2.1] p{0.6em}
                S[table-format=2.1] S[table-format=2.1] p{0.6em}
                S[table-format=2.1] S[table-format=2.1]}
\toprule
& \multicolumn{2}{c}{MedQA} && \multicolumn{2}{c}{MedMCQA} && \multicolumn{2}{c}{BBH} \\
\cmidrule(lr){2-3} \cmidrule(lr){5-6} \cmidrule(lr){8-9}
Policy & {Acc.\,$\uparrow$} & {Lat.\,$\downarrow$}
      && {Acc.\,$\uparrow$} & {Lat.\,$\downarrow$}
      && {Acc.\,$\uparrow$} & {Lat.\,$\downarrow$} \\
\midrule
\multicolumn{9}{l}{\textit{Device}} \\
direct           & 78.8 &  9.8 && 63.6 &  7.8 && 82.2 &  5.7 \\
chain-of-thought & 86.2 & 81.0 && 68.6 & 66.4 && 90.6 & 44.9 \\
self-consistency & 80.0 & 29.1 && 64.6 & 23.2 && 85.2 & 16.8 \\
FrugalGPT        & 85.6 & 70.6 && 63.9 &  8.7 && 83.2 &  8.2 \\
Self-REF         & 84.5 & 45.0 && 66.7 & 33.9 && 85.7 & 12.2 \\
\textbf{HARISSA} & \bfseries 85.7 & \bfseries 33.0 && \bfseries 68.5 & \bfseries 38.4 && \bfseries 88.7 & \bfseries 11.7 \\
\midrule
\multicolumn{9}{l}{\textit{Server}} \\
FrugalGPT        & 69.4 & 11.4 && 59.7 &  9.9 && 71.4 &  6.8 \\
Self-REF         & 69.7 & 11.4 && 59.7 &  9.0 && 71.9 &  7.3 \\
\textbf{HARISSA} & \bfseries 81.8 & \bfseries 11.6 && \bfseries 60.2 & \bfseries 7.1 && \bfseries 91.3 & \bfseries 8.1 \\
\bottomrule
\end{tabular}
\end{table}

\subsection{Single model on device}
\label{sec:ondevice}

\paragraph{HARISSA is within one accuracy point of chain-of-thought at 2.7 times lower latency.}
Chain-of-thought is 7.4 points more accurate than direct on MedQA at about eight times its
latency (Table~\ref{tab:main}), so a policy that thinks only on the queries direct gets wrong
could keep most of that gain at far lower latency. Averaged over the three tasks, HARISSA is 0.8 points below chain-of-thought at 2.7 times lower latency. On MedQA it answers 85.7\% of queries correctly at $33.0$\,s against chain-of-thought's 86.2\% at $81.0$\,s. The gap is widest on BBH, where HARISSA is 1.9 points below chain-of-thought at 3.8 times lower latency. HARISSA is more accurate than self-consistency on every task.

\paragraph{HARISSA reaches FrugalGPT's accuracy on MedQA by thinking on 29\% of queries where FrugalGPT thinks on 75\%.}
Every cascade on the device generates the direct answer first and thinks only on the queries whose direct answer it does not trust, so its latency rises with the share of queries it sends to think, and the better cascade reaches a given accuracy by thinking on fewer queries. On MedQA HARISSA thinks on 29\% of queries and FrugalGPT on 75\% for the same accuracy (Table~\ref{tab:commit}), and Self-REF thinks on 43\% and is 1.2 points less accurate. On BBH, Self-REF and HARISSA think on a similar share of queries (15 and 13\%) and gain 3.5 and 6.5 points over direct. On MedMCQA, FrugalGPT delivers nearly every direct answer and stays at direct's accuracy. HARISSA is at least as accurate as FrugalGPT and Self-REF at every latency its sweep reaches (Figure~\ref{fig:device}, top), because the answer probe finds more of the wrong direct answers than FrugalGPT's scorer or Self-REF's confidence.

\paragraph{On Gemma~4 and Cogito, HARISSA is within one point of chain-of-thought averaged over the three tasks, and thinks more often where thinking adds more accuracy per second of latency.}
How much a cascade can save depends on how much thinking costs and how much accuracy it adds, and the three families we run differ on both. On Gemma~4, think costs 1.5 times direct on MedQA, against 8.3 times on Qwen~3.5, so there is little latency to save; HARISSA thinks on 69\% of MedQA queries (26\% on MedMCQA, 20\% on BBH) and is 0.7 points above chain-of-thought at 1.1 times lower latency, averaged over the three tasks. On Cogito, think costs 3.7 times direct and adds at least 9 points on every task, so on MedQA the policy thinks on every query and matches chain-of-thought exactly, and on BBH it is 0.1 points below at 1.7 times lower latency (Appendix~\ref{app:replication}).

\subsection{Server cascade}
\label{sec:onprem}

\paragraph{At similar latency on MedQA, HARISSA is 12 points more accurate than FrugalGPT and Self-REF.}
On the server the actions are the four sizes of one model. The 2B size answers MedQA in $5.8$\,s and answers half of the queries correctly, and the 27B size takes $29.9$\,s and answers 88\% correctly (Table~\ref{tab:selected}), so a cascade over the sizes saves latency by answering with the smallest size that answers correctly. FrugalGPT and Self-REF generate at the 2B size on every query and deliver its answer on 43\% of MedQA queries, at $11.4$\,s against HARISSA's $11.6$\,s, and across the sweep HARISSA's curve lies above FrugalGPT's and Self-REF's on every task (Figure~\ref{fig:server}, top). On BBH HARISSA is 20 points more accurate than FrugalGPT and Self-REF at about $1$\,s more per query. On MedMCQA HARISSA also delivers the 2B size's answer on a quarter of queries, so the three cascades are equally accurate; when one accuracy point is worth $1$\,s of latency instead of $0.5$\,s, the policy skips the 2B size more often and is 12 points more accurate than FrugalGPT and Self-REF (Table~\ref{tab:price}). Against the 27B size alone, HARISSA is 6.6 points less accurate on MedQA at 2.6 times lower latency.

\subsection{Deferring to the person}
\label{sec:deferral}

\paragraph{At 10\% of queries deferred, the answer probability leaves the fewest wrong answers on five of six task and setting pairs.}
The answer probe already scores every answer HARISSA delivers, so the policy can defer the
least probable answers to the person without generating anything more. The answer probability
and the three standard confidence signals (sequence likelihood, the logit margin, and Self-REF
confidence) each rank the answers HARISSA delivers, and the lowest-ranked share is deferred
(Figures~\ref{fig:device} and~\ref{fig:server}, bottom). At 10\% deferred, the answer probability leaves wrong answers on 4.9\% of device BBH queries, sequence likelihood on 7.6\%, and Self-REF confidence on 7.9\%. The logit margin is undefined for the 13\% of BBH answers whose option letter cannot be located in the generation, and those answers are deferred first, so at 10\% deferred it has deferred only answers without a margin. On server MedMCQA, the one exception, the answer probability
is about one point behind sequence likelihood and the logit margin. The policy defers 8.7\% of MedQA queries on the device and removes 32\% of the wrong answers it would otherwise deliver, nearly four times what deferring the same share at random removes (Table~\ref{tab:deferral-auroc}).

\section{Analysis}
\label{sec:analysis}

Each hidden state predicts whether the answer is correct, and how well it does so decides which probe the policy relies on and whether the probes can choose a size per query.

\begin{figure}[t]
\centering
\includegraphics[width=0.85\textwidth]{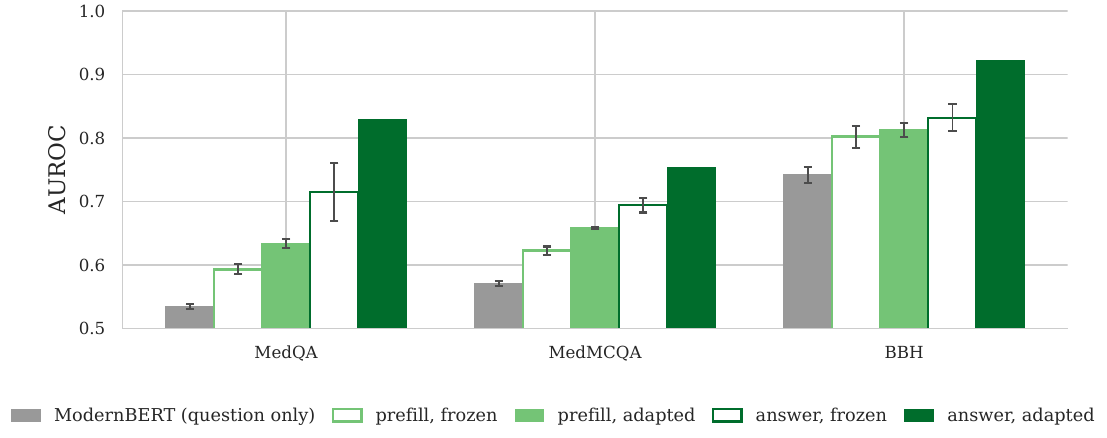}
\caption{\textbf{On the device, the answer probe predicts correctness better than the prefill
probe, and fine-tuning improves the prefill probe and the answer probe.} AUROC on the device,
averaged over the two actions, of a question-text classifier (grey), the prefill probe (light
green), and the answer probe (dark green), frozen (outlined) and fine-tuned (filled); whiskers
are the standard deviation over seeds where more than one seed was run.}
\label{fig:auroc}
\end{figure}

\subsection{What does each hidden state know about the answer?}
\label{sec:analysis-signal}

\paragraph{The answer state predicts correctness better than the prefill state.}
Before decoding, the hidden state reflects only the question, and after decoding it also reflects the answer the model generated, so the answer state can carry a signal the prefill state cannot. We measure how well each probe separates correct from wrong answers by its AUROC, the probability that a randomly chosen correct answer scores higher than a randomly chosen wrong one (Figure~\ref{fig:auroc}). On the device the answer probe reaches 0.83 AUROC on MedQA and the prefill probe 0.63; on BBH the two are closer, 0.92 and 0.81, and every server size shows the same order (Table~\ref{tab:auroc-full}). The prefill probe comes closest on BBH, where a ModernBERT classifier fit on the question text alone reaches 0.74 AUROC against 0.53 on MedQA, so the difficulty on BBH is visible in the question and on the medical tasks depends on whether the model knows the fact. The classifier is below the prefill probe on every task, because a classifier on the question learns which questions are hard in general, while the prefill state is the model's own processing of the question and reflects the question's difficulty and what this model knows.

\paragraph{The cost of the actions decides which probe the policy relies on.}
Skipping an action on the prefill probe's prediction pays when the latency it saves is worth more, at the deployment's price, than the accuracy the wrong skips lose. On the device, think costs about eight times direct
but the prefill probe is weak on the medical tasks, so the development sweep selects a skip threshold of $0$ on every task. The policy therefore generates every direct
answer and lets the answer check decide, and the skip is selected on MedQA only when one accuracy point is worth $30$\,s of latency instead of $10$\,s (Table~\ref{tab:price}). Without the answer check, the policy decides from the prefill
probe alone and is less accurate on every task and slower on MedQA and BBH (85.2\% at $55.1$\,s
against 85.7\% at $33.0$\,s on MedQA; Table~\ref{tab:selected}). On the server a skipped size costs only its prefill pass and probe while a generated size costs its full latency, so the policy skips the 2B size on most queries.
Without the prefill skip it generates at the 2B size on every query, as FrugalGPT and Self-REF
do, and is 6.3 points less accurate on MedQA at similar latency.

\paragraph{Fine-tuning improves the answer probe and barely changes the answers.}
Fine-tuning trains the model to predict correctness at the answer state (Section~\ref{sec:finetuning}), and the answer probe is where its gain lands. On the device it raises the answer probe by 0.12 AUROC on
MedQA (0.06 on MedMCQA, 0.09 on BBH) and the prefill probe by at most 0.04
(Figure~\ref{fig:auroc}), while each action's accuracy moves by at most 1.7 points in either setting (Table~\ref{tab:adapter}). On the device, the policy run on the model without fine-tuning, with the probes refit on the frozen states, is 2.5 points less accurate on MedQA (5.5 on MedMCQA, 2.6 on BBH). On the server the gain is smaller, 1.3 points on MedQA and 1.7 on BBH, and on MedMCQA the frozen policy selects a slower point under the $0.5$\,s price, 7 points more accurate at 1.7 times the latency (Table~\ref{tab:selected}).

\subsection{Can the probes route instead of cascade?}
\label{sec:analysis-route}

\begin{table}[t]
\centering
\small
\caption{\textbf{Routing from the answer probabilities gains about one point over the 27B size at twice its latency.} Test accuracy at mean latency per query on the server; the 27B size on every query is the reference. The prefill router generates at the first size whose prefill probability passes a threshold (HARISSA without the answer check); the answer router generates at every size and delivers the answer with the highest answer probability.}
\label{tab:routing}
\begin{tabular}{lccc}
\toprule
Policy & MedQA & MedMCQA & BBH \\
\midrule
27B alone      & 88.4\% @ 29.9\,s & 74.7\% @ 22.9\,s & 93.3\% @ 15.2\,s \\
\midrule
HARISSA        & 81.8\% @ 11.6\,s & 60.2\% @ 7.1\,s  & 91.3\% @ 8.1\,s \\
prefill router & 77.3\% @ 10.1\,s & 59.3\% @ 7.2\,s  & 88.4\% @ 7.8\,s \\
answer router  & 89.7\% @ 66.0\,s & 73.6\% @ 49.7\,s & 94.4\% @ 36.6\,s \\
\bottomrule
\end{tabular}
\end{table}

\paragraph{Routing from the answer probabilities gains about one point over the 27B size at twice its latency.}
A router chooses one size per query instead of cascading, and the probes give two ways to build one. The prefill router reads each size's prefill probability, cheapest size first, and generates at the first size whose probability passes a threshold, which is HARISSA without the answer check. The answer router generates at every size, reads each answer's probability, and delivers the answer with the highest. Every size answered every test query, and a router that always chose a size whose answer is correct would reach 95.3\% on MedQA (86.3\% on MedMCQA, 95.9\% on BBH), 7 points above the 27B size (12 on MedMCQA, 3 on BBH; Table~\ref{tab:routing}), so routing has room to gain. The prefill router is less accurate than the cascade at similar latency on every task (77.3\% against 81.8\% on MedQA), because it delivers the 4B size's wrong answers that the cascade would pass to the 9B size. The answer router gains 1.3 points over the 27B size on MedQA and 1.1 on BBH, loses 1.1 on MedMCQA, and costs about twice the 27B size's latency. Each probe is calibrated within one size, so its output estimates whether that size's answer is correct, and comparing two such outputs does not pick the better size. The policy therefore cascades, and accuracy above the 27B size requires a probe that compares sizes.

\section{Conclusion}
\label{sec:conclusion}

We show that a small model's own hidden states can decide, for each query, whether to answer directly or reason first, which size of the model to run, and whether to deliver the answer or defer the query to the person. A probe on the prefill state skips an action the model is predicted to get wrong before it generates, and a probe on the answer state withholds an answer predicted wrong. On a device, HARISSA is within one accuracy point of chain-of-thought at 2.7 times lower latency, averaged over three tasks. On a server holding four sizes of one model, HARISSA is more accurate than FrugalGPT and Self-REF at the same latency, and at the same deferral rate the answer probability leaves fewer wrong answers than the standard confidence signals in five of six task and setting pairs.

\section*{Reproducibility Statement}
The fine-tuning, probe, and cascade code, the configuration of every run, and the graded answers behind every table are in the supplementary material and will be released on publication. Section~\ref{sec:setting} states the objective, Algorithm~\ref{alg:policy} the policy, and Section~\ref{sec:finetuning} and Appendix~\ref{app:training} the fine-tuning recipe, the probed layers, and the probe and calibration settings. Appendix~\ref{app:data} gives the split sizes, the BBH subtasks, and the prompt and decoding settings; Section~\ref{sec:tasks} and Appendix~\ref{app:replication} name every model, with Hugging Face identifiers for Gemma~4 and Cogito. Latency is computed from per-size prefill and per-token rates measured on one NVIDIA A40, listed with the generated-token counts in Appendix~\ref{app:latency}, so every latency in the paper can be recomputed from the released outputs. Section~\ref{sec:metrics} fixes the splits on which the probes are fit, the thresholds chosen, and the results reported, and Appendices~\ref{app:results} to~\ref{app:finetuning} report every policy, ablation, size, deferral signal, and probe on the test split.

\section*{AI use statement}
In this work, we used generative AI tools to draft and edit the paper's text, to review its
structure and prose, to help implement the method and the results-reporting and figure
scripts, and, in part, to analyze data, to interpret results, and to design experiments. We
have not used generative AI tools to generate data, to formulate or prove claims, or to clean
datasets. Every number in the paper was checked against the tables printed by our own code,
and all AI-assisted text was reviewed by the authors. We take responsibility for the final
content of this work, including text, claims, or artifacts produced with the aid of generative
AI.

\bibliography{iclr2026_conference}
\bibliographystyle{iclr2027_conference}

\clearpage
\appendix
\raggedbottom
\renewcommand{\floatpagefraction}{0.75}
\renewcommand{\bottomfraction}{0.9}
\setcounter{totalnumber}{4}
\setcounter{topnumber}{3}
\setcounter{bottomnumber}{3}
\renewcommand{\textfraction}{0.05}

\section{Datasets and prompts}
\label{app:data}

Table~\ref{tab:datasets} gives the split sizes. From BBH we use the sixteen multiple-choice subtasks:
tracking\_shuffled\_objects\_\{three,five,seven\}\_objects,
logical\_deduction\_\{three,five,seven\}\_objects, disambiguation\_qa,
salient\_translation\_error\_detection, date\_understanding, hyperbaton, temporal\_sequences,
movie\_recommendation, ruin\_names, snarks, penguins\_in\_a\_table, and
reasoning\_about\_colored\_objects.

\begin{table}[H]
\centering
\small
\caption{Split sizes (queries).}
\label{tab:datasets}
\begin{tabular}{@{}lrrr@{}}
\toprule
Task & Train & Dev & Test \\
\midrule
MedQA   & 10{,}178 & 1{,}272 & 1{,}273 \\
MedMCQA & 10{,}178 & 2{,}091 & 2{,}092 \\
BBH     & 2{,}188  & 552     & 906 \\
\bottomrule
\end{tabular}
\end{table}

\paragraph{Prompt and decoding.} Every action uses the zero-shot prompt below, with one
reasoning hint per task (for the medical tasks, ``step-by-step analysis of the medical scenario
and answer choices''). On the device, direct and think differ only in the thinking flag of the
chat template, so the prefill state is identical up to the template's mode tokens. Under think,
the model's thinking precedes the JSON. Grading reads the \texttt{final\_answer} letter. Every
model decodes at temperature $0.6$ with a $10{,}000$-token cap on the response, thinking
included.

\begin{quote}\small\ttfamily
\# Question\\
\{question text, with lettered options\}\\[4pt]
\# Required Output Format\\
```json\\
\{\\
\hspace*{1em}"reasoning": "<\{reasoning hint\}>",\\
\hspace*{1em}"final\_answer": "A"\\
\}\\
```
\end{quote}

\section{Fine-tuning and probe details}
\label{app:training}
We fine-tune a LoRA adapter on the attention projections $\{q,k,v,o\}$ (rank $16$,
$\alpha=32$, dropout $0.05$) with batch size $8$, learning rate $10^{-4}$, and sequence length
$3072$, for three epochs (one epoch for the 27B size). The same recipe is used for the device
model and for each server size, which is fine-tuned separately; Cogito uses learning rate
$2\times10^{-5}$. Training labels grade the frozen model's stored answers, and one forward pass
over the prompt and that answer trains the answer head on the last generated token and one
prefill head per action on the last prompt token, under a binary cross-entropy loss at equal
weight. The development split is evaluated every $250$ steps; on the device we keep the
checkpoint with the lowest cascade objective on the development split, and on the server the
checkpoint whose answer head best separates correct from incorrect answers there. The heads are
discarded, and the probes are fit on the fine-tuned model's own answers.

The probed layer is fixed per size: layer $32$ for the 4B and 9B sizes, layer $64$ for the 27B
size, and the final layer for the 2B size, chosen once from a layer sweep on MedQA; Gemma~4 is
probed at its final layer and Cogito at layer $32$. Each probe is an L2-regularized logistic
regression ($C=0.5$, L-BFGS) on that layer's hidden state, with features standardized by
training-split statistics, fit on the training split and calibrated by isotonic regression on
the development split, with the development scores cross-fitted and out-of-range scores clipped.

\section{Cost model}
\label{app:latency}
A query's latency is a per-size prefill time plus its generated tokens times a per-token
generation rate (Table~\ref{tab:rates}). Both were measured for each size on one NVIDIA A40
(48\,GB, vLLM, BF16, concurrency one) on eight MedQA development prompts, and the same profile
serves both settings, so the device numbers are those of a 4B model on that GPU rather than on
a phone. Table~\ref{tab:tokens} gives the mean generated tokens per action and the resulting
latency of each action alone. Each probe costs one dot product on a state the model computes
while answering. A skipped action costs its prefill pass and probe: $0.08$\,s for direct on
the device, and $0.23$\,s for the 2B, 4B, and 9B sizes together on the server, measured as the
latency of the policy at a skip threshold above $1$, which skips every action but the last,
minus the latency of the last action alone.

\begin{table}[H]
\centering
\small
\caption{\textbf{Measured prefill time and per-token generation time on one A40.}}
\label{tab:rates}
\begin{tabular}{@{}lrr@{}}
\toprule
Model & Prefill (ms) & Generation (ms per token) \\
\midrule
Qwen~3.5 2B   & 44.0  & 7.92 \\
Qwen~3.5 4B   & 82.5  & 17.14 \\
Qwen~3.5 9B   & 104.0 & 30.45 \\
Qwen~3.5 27B  & 222.8 & 49.56 \\
Gemma~4 E4B   & 64.5  & 19.52 \\
Cogito v1 8B  & 87.2  & 27.98 \\
\bottomrule
\end{tabular}
\end{table}

\begin{table}[H]
\centering
\small
\setlength{\tabcolsep}{4pt}
\caption{\textbf{Mean generated tokens per query and the latency of each action alone (s), Qwen~3.5, test split.} Input is the mean prompt length in tokens; the 4B size is the device model's direct action, so the two columns agree.}
\label{tab:tokens}
\begin{tabular}{@{}lrrrrrrr@{}}
\toprule
 & & \multicolumn{6}{c}{Generated tokens} \\
\cmidrule(lr){3-8}
Task & Input & direct & think & 2B & 4B & 9B & 27B \\
\midrule
MedQA   & 225 & 565 & 4{,}723 & 725 & 565 & 670 & 599 \\
MedMCQA & 46  & 450 & 3{,}868 & 515 & 450 & 485 & 457 \\
BBH     & 113 & 325 & 2{,}614 & 737 & 325 & 317 & 303 \\
\midrule
 & & \multicolumn{6}{c}{Latency of the action alone (s)} \\
\cmidrule(lr){3-8}
MedQA   & & 9.8 & 81.0 & 5.8 & 9.8 & 20.5 & 29.9 \\
MedMCQA & & 7.8 & 66.4 & 4.1 & 7.8 & 14.9 & 22.9 \\
BBH     & & 5.7 & 44.9 & 5.9 & 5.7 & 9.8  & 15.2 \\
\bottomrule
\end{tabular}
\end{table}

\section{Full results in both settings}
\label{app:results}

\begin{table}[H]
\centering
\footnotesize
\setlength{\tabcolsep}{4pt}
\caption{\textbf{Ablations, each size alone, and the policy without fine-tuning: accuracy (\% of queries) at mean latency per query (s) at the selected thresholds.} Thresholds are chosen on the development split at $10$\,s per accuracy point on the device and $0.5$\,s on the server; HARISSA and the baselines are in Table~\ref{tab:main}, and the server policy without the answer check is the prefill router of Table~\ref{tab:routing}. HARISSA without fine-tuning runs the policy on the frozen model with probes fit on its states.}
\label{tab:selected}
\begin{tabular}{lccc}
\toprule
Policy & MedQA & MedMCQA & BBH \\
\midrule
\textit{Device} & & & \\
HARISSA without the prefill skip & 86.6\% @ 39.7\,s & 67.5\% @ 26.5\,s & 88.6\% @ 11.6\,s \\
HARISSA without the answer check & 85.2\% @ 55.1\,s & 66.2\% @ 27.0\,s & 86.5\% @ 15.0\,s \\
HARISSA without fine-tuning & 83.2\% @ 36.5\,s & 63.0\% @ 7.9\,s & 86.1\% @ 13.8\,s \\
\midrule
\textit{Server} & & & \\
2B alone & 45.6\% @ 5.8\,s & 47.1\% @ 4.1\,s & 67.9\% @ 5.9\,s \\
4B alone & 78.8\% @ 9.8\,s & 63.1\% @ 7.8\,s & 82.8\% @ 5.7\,s \\
9B alone & 83.3\% @ 20.5\,s & 70.1\% @ 14.9\,s & 87.3\% @ 9.8\,s \\
27B alone & 88.4\% @ 29.9\,s & 74.7\% @ 22.9\,s & 93.3\% @ 15.2\,s \\
HARISSA without the prefill skip & 75.5\% @ 13.3\,s & 58.1\% @ 8.0\,s & 89.1\% @ 11.8\,s \\
HARISSA without the answer check & 77.3\% @ 10.1\,s & 59.3\% @ 7.2\,s & 88.4\% @ 7.8\,s \\
HARISSA without fine-tuning & 80.5\% @ 12.4\,s & 67.2\% @ 11.9\,s & 89.6\% @ 8.4\,s \\
\bottomrule
\end{tabular}
\end{table}

\begin{table}[H]
\centering
\footnotesize
\setlength{\tabcolsep}{3pt}
\caption{\textbf{The action each cascade delivers from (\% of queries), at the selected thresholds.}}
\label{tab:commit}
\begin{tabular}{llcc@{\hspace{1.6em}}cccc}
\toprule
 & & \multicolumn{2}{c}{Device} & \multicolumn{4}{c}{Server} \\
\cmidrule(lr){3-4}\cmidrule(lr){5-8}
Task & Policy & direct & think & 2B & 4B & 9B & 27B \\
\midrule
MedQA   & HARISSA   & 71.4 & 28.6 & 9.0  & 78.9 & 12.1 & 0.0 \\
        & FrugalGPT & 25.0 & 75.0 & 43.3 & 56.4 & 0.3  & 0.0 \\
        & Self-REF  & 56.6 & 43.4 & 43.1 & 56.8 & 0.1  & 0.0 \\
MedMCQA & HARISSA   & 54.0 & 46.0 & 27.2 & 71.0 & 1.8  & 0.0 \\
        & FrugalGPT & 98.7 & 1.3  & 27.9 & 70.8 & 1.3  & 0.0 \\
        & Self-REF  & 60.7 & 39.3 & 36.9 & 63.1 & 0.0  & 0.0 \\
BBH     & HARISSA   & 86.6 & 13.4 & 0.0  & 68.4 & 18.7 & 12.9 \\
        & FrugalGPT & 94.3 & 5.7  & 89.3 & 7.3  & 3.4  & 0.0 \\
        & Self-REF  & 85.3 & 14.7 & 86.5 & 6.5  & 7.0  & 0.0 \\
\bottomrule
\end{tabular}
\end{table}

\begin{table}[H]
\centering
\footnotesize
\setlength{\tabcolsep}{3pt}
\caption{\textbf{HARISSA's selected point at each price on latency: accuracy at mean latency
per query, with the selected skip threshold in parentheses.} The paper reports $10$\,s per
accuracy point on the device and $0.5$\,s on the server.}
\label{tab:price}
\begin{tabular}{lcccc}
\toprule
Task & \multicolumn{4}{c}{Seconds per accuracy point} \\
\midrule
\textit{Device} & 2 & 5 & 10 & 30 \\
\cmidrule(lr){2-5}
MedQA   & 82.6\% @ 18.2\,s (0.00) & 83.3\% @ 19.5\,s (0.00) & 85.7\% @ 33.0\,s (0.00) & 86.9\% @ 49.8\,s (0.73) \\
MedMCQA & 63.6\% @ 7.9\,s (0.00)  & 63.6\% @ 7.9\,s (0.00)  & 68.5\% @ 38.4\,s (0.00) & 68.5\% @ 38.5\,s (0.03) \\
BBH     & 88.7\% @ 11.7\,s (0.00) & 88.7\% @ 11.7\,s (0.00) & 88.7\% @ 11.7\,s (0.00) & 88.7\% @ 11.7\,s (0.00) \\
\midrule
\textit{Server} & 0.1 & 0.25 & 0.5 & 1 \\
\cmidrule(lr){2-5}
MedQA   & 68.2\% @ 10.1\,s (0.40) & 81.8\% @ 11.6\,s (0.60) & 81.8\% @ 11.6\,s (0.60) & 84.4\% @ 13.5\,s (0.60) \\
MedMCQA & 53.6\% @ 5.7\,s (0.00)  & 60.2\% @ 7.1\,s (0.45)  & 60.2\% @ 7.1\,s (0.48)  & 72.6\% @ 16.1\,s (0.73) \\
BBH     & 85.0\% @ 7.2\,s (0.80)  & 90.4\% @ 7.9\,s (0.88)  & 91.3\% @ 8.1\,s (0.88)  & 91.9\% @ 8.4\,s (0.88) \\
\bottomrule
\end{tabular}
\end{table}

\section{Replication on other model families}
\label{app:replication}

The device experiments of Section~\ref{sec:ondevice} are repeated on Gemma~4 E4B
\citep{gemma4TechnicalReport2026} and Cogito v1 8B \citep{cogitoV1PreviewModelCard2025} with
the same tasks, splits, actions, fine-tuning recipe, probes, threshold selection at $10$\,s per
accuracy point, and cost model, each family with its own measured rates and token counts
(Appendix~\ref{app:latency}). Table~\ref{tab:replication} reports each policy at its selected
thresholds, the share of queries HARISSA thinks on, and each probe's AUROC; Section~\ref{sec:ondevice} reads the table.

\begin{table}[H]
\centering
\footnotesize
\setlength{\tabcolsep}{4pt}
\caption{\textbf{Gemma~4 and Cogito on the device.} Accuracy (\% of queries) at mean latency
per query (s) for each policy at its selected thresholds, the share of queries HARISSA thinks on (a dash marks a share not recorded), and the AUROC of the fine-tuned probes averaged over the two actions;
test split, one seed.}
\label{tab:replication}
\setlength{\tabcolsep}{2.5pt}
\begin{tabular}{@{}llcccccc@{}}
\toprule
 & & \multicolumn{3}{c}{Accuracy @ latency} & HARISSA & \multicolumn{2}{c}{Probe AUROC} \\
\cmidrule(lr){3-5}\cmidrule(lr){6-6}\cmidrule(lr){7-8}
Family & Task & direct & chain-of-thought & HARISSA & thinks on & prefill & answer \\
\midrule
Gemma~4 & MedQA   & 66.1\% @ 17.0\,s & 74.1\% @ 24.8\,s & 73.9\% @ 25.7\,s & 69.4\% & 0.611 & 0.766 \\
        & MedMCQA & 56.8\% @ 13.9\,s & 58.7\% @ 20.2\,s & 59.5\% @ 19.2\,s & 26.1\% & 0.658 & 0.759 \\
        & BBH     & 80.9\% @ 7.6\,s  & 82.7\% @ 14.4\,s & 84.1\% @ 10.5\,s & 19.5\% & 0.878 & 0.909 \\
\midrule
Cogito  & MedQA   & 58.4\% @ 9.6\,s  & 67.6\% @ 35.7\,s & 67.6\% @ 35.7\,s & 100.0\% & 0.611 & 0.727 \\
        & MedMCQA & 48.1\% @ 6.7\,s  & 61.6\% @ 25.0\,s & 61.0\% @ 23.2\,s & -- & 0.645 & 0.676 \\
        & BBH     & 76.8\% @ 4.6\,s  & 86.4\% @ 22.6\,s & 86.3\% @ 13.0\,s & -- & 0.753 & 0.841 \\
\bottomrule
\end{tabular}
\end{table}

\section{Deferral}
\label{app:deferral}

Table~\ref{tab:deferral-auroc} gives HARISSA's deferral in both settings when the deferral threshold equals the selected fall-through threshold, so that the answer probability that discards an answer at a non-last action also withholds it at the last.

\begin{table}[H]
\centering
\footnotesize
\setlength{\tabcolsep}{2.5pt}
\caption{\textbf{Deferral by the answer probability at the selected fall-through threshold, on HARISSA's delivered answers at its selected thresholds.} The share of queries deferred, and the wrong answers delivered (\% of queries) with no deferral and at that threshold.}
\label{tab:deferral-auroc}
\begin{tabular}{lccc}
\toprule
 & & \multicolumn{2}{c}{Wrong answers delivered} \\
\cmidrule(lr){3-4}
Task & Deferred & none & at the threshold \\
\midrule
\textit{Device} & & & \\
MedQA   & 8.7\%  & 14.3 & 9.7  \\
MedMCQA & 23.7\% & 31.5 & 17.4 \\
BBH     & 4.2\%  & 11.3 & 7.4  \\
\midrule
\textit{Server} & & & \\
MedQA   & 0.0\%  & 18.2 & 18.2 \\
MedMCQA & 0.0\%  & 39.8 & 39.8 \\
BBH     & 2.8\%  & 8.7  & 6.0  \\
\bottomrule
\end{tabular}
\end{table}

\section{Fine-tuning's effect on the probes and the answers}
\label{app:finetuning}

Table~\ref{tab:auroc-full} gives each probe's AUROC on the frozen and the fine-tuned model
beside a classifier on the question text, and Table~\ref{tab:adapter} gives each action's
accuracy on both models.

\begin{table}[H]
\centering
\footnotesize
\caption{\textbf{Per-action AUROC of each probe, test split.} On the device, frozen and fine-tuned, beside a ModernBERT classifier fit on the question text; on the server, the fine-tuned probes of each size; one seed per cell.}
\label{tab:auroc-full}
\begin{tabular}{llccccc}
\toprule
 & & & \multicolumn{2}{c}{Prefill probe} & \multicolumn{2}{c}{Answer probe} \\
\cmidrule(lr){4-5}\cmidrule(lr){6-7}
Task & Action & Question only & frozen & fine-tuned & frozen & fine-tuned \\
\midrule
MedQA   & direct & 0.540 & 0.594 & 0.639 & 0.716 & 0.840 \\
        & think  & 0.539 & 0.572 & 0.610 & 0.812 & 0.818 \\
MedMCQA & direct & 0.573 & 0.611 & 0.655 & 0.680 & 0.759 \\
        & think  & 0.574 & 0.645 & 0.659 & 0.716 & 0.746 \\
BBH     & direct & 0.714 & 0.735 & 0.762 & 0.752 & 0.914 \\
        & think  & 0.805 & 0.828 & 0.855 & 0.852 & 0.928 \\
\midrule
MedQA   & 2B  & & & 0.604 & & 0.777 \\
        & 4B  & & & 0.679 & & 0.836 \\
        & 9B  & & & 0.650 & & 0.819 \\
        & 27B & & & 0.745 & & 0.851 \\
MedMCQA & 2B  & & & 0.615 & & 0.720 \\
        & 4B  & & & 0.702 & & 0.791 \\
        & 9B  & & & 0.681 & & 0.738 \\
        & 27B & & & 0.727 & & 0.780 \\
BBH     & 2B  & & & 0.702 & & 0.822 \\
        & 4B  & & & 0.785 & & 0.897 \\
        & 9B  & & & 0.824 & & 0.875 \\
        & 27B & & & 0.830 & & 0.881 \\
\bottomrule
\end{tabular}
\end{table}

\begin{table}[H]
\centering
\footnotesize
\caption{\textbf{Fine-tuning changes the answers by at most 1.7 points.} Accuracy (\% of
queries) of each fixed action on the frozen and the fine-tuned model, as frozen$\to$fine-tuned
($\Delta$); test split, one seed.}
\label{tab:adapter}
\begin{tabular}{lcccc}
\toprule
Task & \multicolumn{4}{c}{Action} \\
\midrule
\textit{Device} & direct & think & & \\
\cmidrule(lr){2-3}
MedQA   & 78.6$\to$78.8 ($+0.2$) & 84.8$\to$86.2 ($+1.3$) & & \\
MedMCQA & 63.0$\to$63.6 ($+0.7$) & 66.9$\to$68.6 ($+1.7$) & & \\
BBH     & 83.8$\to$82.2 ($-1.5$) & 89.1$\to$90.6 ($+1.5$) & & \\
\midrule
\textit{Server} & 2B & 4B & 9B & 27B \\
\cmidrule(lr){2-5}
MedQA   & 45.6$\to$45.6 ($+0.1$) & 78.6$\to$78.8 ($+0.2$) & 82.9$\to$83.3 ($+0.5$) & 88.8$\to$88.4 ($-0.5$) \\
MedMCQA & 46.5$\to$47.1 ($+0.6$) & 63.0$\to$63.1 ($+0.1$) & 70.2$\to$70.1 ($-0.1$) & 75.2$\to$74.7 ($-0.5$) \\
BBH     & 68.9$\to$67.9 ($-1.0$) & 83.8$\to$82.8 ($-1.0$) & 86.4$\to$87.3 ($+0.9$) & 93.3$\to$93.3 ($+0.0$) \\
\bottomrule
\end{tabular}
\end{table}

\end{document}

%% file: math_commands.tex
\usepackage{amsmath,amsfonts,bm}

\def\eqref#1{equation~\ref{#1}}

\def\1{\bm{1}}

\DeclareMathAlphabet{\mathsfit}{\encodingdefault}{\sfdefault}{m}{sl}
\SetMathAlphabet{\mathsfit}{bold}{\encodingdefault}{\sfdefault}{bx}{n}